\documentclass{article}

\usepackage[preprint]{corl_2021} 

\usepackage{amsmath}
\usepackage{amssymb}
\usepackage{todonotes}
\usepackage{wrapfig}

\newcommand{\CnIFull}[0]{\emph{Collect \& Infer}}
\newcommand{\CnI}[0]{\emph{C\&I}}

\newcommand{\InfOp}{\cal{I}}
\newcommand{\CollectOp}{\cal{C}}

\title{Collect \& Infer - a fresh look at data-efficient Reinforcement Learning}

\RequirePackage{graphicx}

\author{
  Martin Riedmiller\\
  DeepMind, UK\\
  \And
    Jost Tobias Springenberg\\
  DeepMind, UK\\
  \And
   Roland Hafner\\
  DeepMind, UK\\
   \And
    Nicolas Heess\\
  DeepMind, UK\\
}

\begin{document}
\maketitle


\begin{abstract}
This position paper proposes a fresh look at Reinforcement Learning (RL) from the perspective of data-efficiency. Data-efficient RL has gone through three major stages: pure on-line RL where every data-point is considered only once, RL with a replay buffer where additional learning is done on a portion of the experience, and finally transition memory based RL, where, conceptually, all transitions are stored and re-used in every update step. While inferring knowledge from all explicitly stored experience has lead to a tremendous gain in data-efficiency, the question of how this data is collected has been vastly understudied. 
We argue that data-efficiency can only be achieved through careful consideration of both aspects. We propose to make this insight explicit via a paradigm that we call 'Collect and Infer', which explicitly models RL as two separate but interconnected processes, concerned with data collection and knowledge inference respectively. We discuss implications of the paradigm, how its ideas are reflected in the literature, and how it can guide future research into data efficient RL. 
\footnote{This paper is based on a keynote talk of the first author at EWRL 2018.}
\end{abstract}


\section{Introduction}

\begin{wrapfigure}{r}{0.3\textwidth}
\vspace{-15pt}
	\includegraphics[width=0.4 \columnwidth]{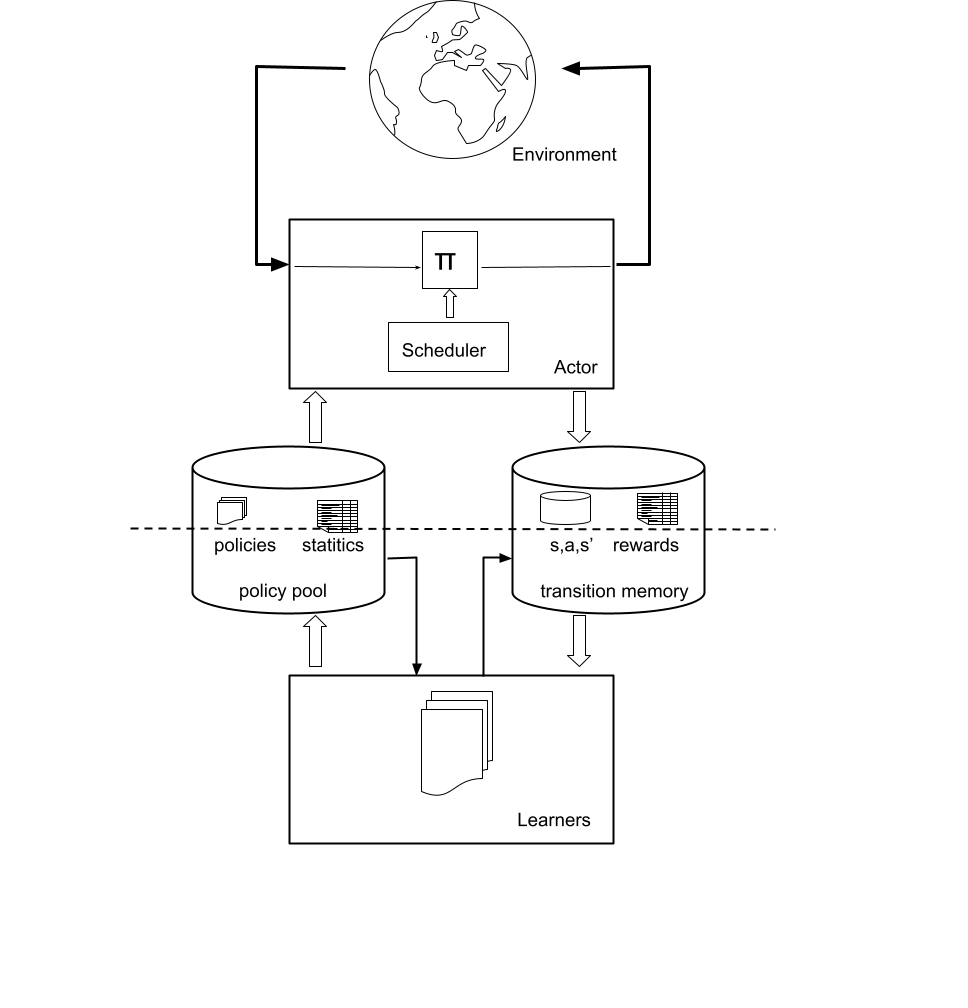}
\vspace{-24pt}	\caption{Collect and Infer Agent. Top part: collecting experience. Lower part: inference. The two parts share policy pool and transition memory.}
	\label{fig:schematic}
	\vspace{-0.5cm}
\end{wrapfigure}

Data-efficiency in Reinforcement Learning (RL) can be loosely characterized as 'getting the most out of the collected experience'. Data-efficiency is critical in many real-world scenarios \cite{dulacarnold2019challenges}, where gathering data is the main bottleneck (e.g. in robotics), but it is also, arguably, a key property of Artificial General Intelligence (AGI).

From a data-efficiency perspective, reinforcement learning methods have gone through three major stages. The original RL framework was phrased in a pure 'online' setting: the agent acts, observes the reward and new state, updates its behaviour and acts again. This view continues to be successful in settings where data is cheap, e.g. if a simulator of the environment is available. The next stage was to introduce a replay buffer \cite{Lin92}, which stored
a subset of the transitions to enhance the learning signal by iterating over recent  experience multiple times. Building on previous work \cite{Boyan99,Lagoudakis03}, \citet{Ernst2005} and \citet{Riedmiller2005} independently suggested to take this idea to the extreme, store all experience in a transition memory and re-use the full data in every update step. This led to a breakthrough in data-efficiency and made the application of model-free RL in the real world possible \cite{RiedmillerFBIT,RHLL08}. 
Recent years have witnessed a revival of this idea with off-policy actor-critic algorithms rapidly gaining importance \cite{heess2015learning,lillicrap2016continuous,abdolmaleki2018maximum,haarnoja2019soft}, especially for robotics applications. 
In parallel there has been a growing interest in RL algorithms that can learn  from fixed data sets entirely without interaction (\emph{offline RL}) \citep{lange2012batch,levineoffline,fujimoto2018offpolicy,siegel2020keep,Gulcehre_2020,wang2020critic,Nair_2020}. 
Together, these lines of work are pointing towards agent designs in which acting and learning are only loosely coupled. However, while the first group of algorithms still lacks a clear separation of data collection and learning, 
in the second group questions like how to work with growing datasets, or how to compose datasets that are effective for offline learning remain largely unstudied.

We extrapolate from these developments and argue that a clear conceptual separation of the reinforcement learning process into two distinct sub-processes, data-collection and inference of knowledge, will lead to further improvements in data efficiency and enhanced capabilities for the next generation of RL agents. We refer to this perspective as the \emph{Collect and Infer} (\CnI)~paradigm. It assumes two sub-processes: acting (data \emph{collection}), and learning (\emph{inference}) which are decoupled but connected through a transition memory into which all data resulting from environment interaction is collected, and from which data is drawn for learning. A particular emphasis is put on how the data is collected.
This view of RL as two independent processes provides additional flexibility in algorithm design and emphasizes that these processes can and should be optimized independently.

This paper gives a light-weight overview of the core concepts and implications of the \CnI~ paradigm. We discuss recent examples from the literature, how these algorithms can be interpreted from the \CnI~perspective, and where that perspective suggests changes or improvements. We conclude with a discussion of research questions motivated by the paradigm. 

\section{The Collect and Infer paradigm and its implications \label{paradigm}}
The key idea of the \CnI~paradigm is to separate Reinforcement Learning into two distinct but interconnected processes: process 1 deals with collecting data into a transition memory by interacting with the environment, process 2 infers knowledge about the environment by learning from the data of said memory. This perspective provides us with a new handle on the question of data efficiency which we can optimize by considering each process separately via the following objectives:

\begin{enumerate}
\item (O1) Given a fixed batch of data, what is the right learning setup, to get to the maximally performing policy (optimal 'inference')?
\item (O2) Given an 'inference' process, what is the minimal set of data, to get to a maximally performing policy (optimal 'collection')?
\end{enumerate}

The \CnI~ perspective has several implications. While it does not prescribe a particular algorithmic solution it encourages us to develop algorithms that satisfy the following desiderata:

\begin{enumerate}
\item Learning is done offline in a 'batch' setting assuming fixed data as suggested by O1. Data may have been collected by a behavior policy different from the one that is the learning target \citep[e.g.][]{riedmiller2018learning}. This enables utilization of the same data to optimize for multiple objectives simultaneously, and coincides with interest in offline RL \citep{lange2012batch,siegel2020keep,wang2020critic,fujimoto2018offpolicy,Nair_2020}.
\item 
Data-collection is a process that should be optimized in its own right. Naive exploration schemes that employ simple random perturbations of a task policy, such as epsilon greedy, are likely to be inadequate. The behavior that is optimal for data collection in the sense of O2 may be quite different from the optimal behavior for a task of interest. 
\item Treating data-collection as a separate process offers novel ways to integrate known methods like
skills, model-based approaches, or innovative exploration schemes into the learning process without
biasing the final task solution.
\item Data collection may happen concurrently with inference (in which case the two processes actively influence each other and we get close to online RL) or can be conducted separately.
\item 
\CnI{} suggests a different focus for evaluation: in contrast to usual regret-based frameworks for exploration, \CnI~ does not aim to optimize task performance during collection.
Instead, we distinguish between a learning phase, during which a certain amount of data is collected, and a deployment phase,
during which the performance of the agent is assessed.
\end{enumerate}

Collect and Infer has implications for agent architectures, and it suggests alternative solutions to a number of problems that will become prominent as RL is applied to more challenging scenarios, including multi-task, transfer or life-long learning \cite{Horde2011}.
The Scheduled Auxiliary Control (SAC-X) architecture of \citet{riedmiller2018learning} exemplifies several of the above ideas.
Its components are an actor, a transition memory, one or more learners, a pool of candidate policies and a scheduler, that selects policies for execution by the actor such as to collect experience that is informative for learning one or multiple tasks. 
Although the SAC-X agent does not explicitly optimize O1 and O2, it does satisfy several of the above desiderata insofar as it decouples data collection and learning and optimizes data collection actively and separately from the task solutions. This is achieved as follows:
(a) The agent optimizes for several auxiliary objectives in parallel to the policies for the primary tasks of interest. (b) This allows the agent to learn a set of auxiliary policies that can facilitate learning of one or more main tasks. (c) These auxiliary policies are deployed to collect better experience. (d) Knowledge is shared across tasks by sharing experience. (e) Execution of auxiliary policies is actively scheduled to improve data collection for the main task. (f) This process is optimized via a separate learning process. The use of auxiliary policies bears some similarity to the role of skills in hierarchical architectures but there are two important differences: (1) Unlike skills, auxiliary policies are not directly used as part of the solution for the main task. The task policy is learned off-policy from the data collected with the auxiliary policies. (2) Execution of the auxiliary policies is scheduled to improve data collection. Although SAC-X emphasizes knowledge sharing via data, as discussed in  e.g.\ \citep{wulfmeier2020compositional}, this can be flexibly combined with a direct reuse of learned behavior representations such as skills.

\section{A formal look at Collect and Infer}
\label{formal}

We provide a partial formalization of the ideas introduced in Section \ref{paradigm}.
We consider the standard objective
consisting of an agent characterized by policy $\pi(a | s)$ acting in an environment $\mathcal{E}$ with states $s \in \mathcal{S}$, actions $a \in \mathcal{A}$, transition probability distribution $p(s_{t+1} | s_t, a_t)$, initial state distribution $p(s_0)$, and reward function $r$. The goal is to find a policy maximizing the expected sum of rewards
\begin{equation}
\textstyle
J(\pi_\theta) = \mathbb{E}_{\tau \sim \pi_\theta} \left[\sum_{t=0}^{T} r(s_t) \right],
\end{equation}
where $\tau = [(s_0, a_0), (s_1, a_1), \dots]$ is a trajectory of length $T$ sampled according to $p$ and $\pi$. 

The main perspective change of \CnI{} is that inference of the policy happens through optimization of a 'surrogate objective' defined in terms of a finite set of data. As a result, the optimization of the data set itself becomes part of the learning process. 
Thus, \CnI{} can be characterized in terms of two operators: a) an 'Inference' operator, $\InfOp$, that given a data set $\mathcal{D}$, computes
a policy $\pi_\theta(a | s)$  and b) a data generation operator, ${\CollectOp}$, 
that generates the data set $\mathcal{D}$. 

More precisely, the collection operator ${\CollectOp}$ will generate a data set consisting of $N$ trajectories, for instance by executing a collection policy $\mu$ in $\mathcal{E}$: $\mathcal{D}_c = \lbrace \tau^1, \dots, \tau^N | \tau^i \sim \mu,p \rbrace = {\CollectOp}(\mathcal{E}, N)$.
The inference operator $\InfOp$ 
optimizes $\pi$ to find the maximum of the surrogate objective $\mathcal{L}_I$ which is defined in terms of data $\mathcal{D}_c$: $
    \pi_\theta = {\InfOp}(\mathcal{D}_c) = \arg \max_{\pi_\theta} \mathcal{L}_I(\pi_\theta, \mathcal{D}_c)$, 
This allows us to express a joint objective that couples  (O1) -- identifying an optimal policy given fixed data -- and (O2) -- identifying an optimal collection process given an inference procedure:
\begin{equation}
\textstyle
    \mathcal{O}({\CollectOp}; {\InfOp}, N) = J(\arg \max_{\pi_\theta} \mathcal{L}_{\mathcal{I}} (\pi_\theta, \mathcal{D}_c = {\CollectOp}(\mathcal{E}, N))).
    \label{eq:collect_obj} 
\end{equation}
We measure the success of the policy inferred from the data collected by $\CollectOp$. For any choice of $\InfOp$, environment $\mathcal{E}$, and fixed data budget $N$ we can identify an optimal collection process via the 'outer' optimization
$
{\CollectOp}^* = \arg\max_{\CollectOp} \mathcal{O}({\InfOp}, {\CollectOp}, N)
$, for instance by optimizing $\mu$. 
Different choices for $\mathcal{I}$ 
will lead to different algorithms with different requirements for $\CollectOp$. 
For instance, we can obtain an algorithm in which we first create a fixed dataset and then obtain the policy via offline RL.

In practice, in particular the optimization with respect to $\CollectOp$ may be intractable and heuristics may be used instead. Furthermore, the collection process and the inference process may be tightly coupled and proceed in an iterative scheme. For example,
the collection process might depend on previous estimates of an optimal policy (or previous data).

\section{\CnIFull~and the state-of-the art in reinforcement learning}
\label{sec:examples}
The example in Section \ref{paradigm}  highlights that the \CnI~ paradigm offers considerable flexibility. It suggests an interpolation between pure offline (batch) and more conventional online learning scenarios, and thus chimes naturally with the growing interest in data driven approaches, where  large datasets of experience are built up over time, which can then enable rapid learning of new behaviors with only small amounts of online experience. Decoupling acting and learning, and the emphasis on off-policy learning gives greater flexibility when designing exploration or other actively optimized data collection strategies, including schemes for unsupervised RL and unsupervised skill discovery. Considering data as a vehicle for knowledge transfer enables new algorithms for multi-task and transfer scenarios. It finally suggests a different emphasis when thinking about meta-learning or life-long learning scenarios. To enable rapid adaptation to a novel task we may, for instance, focus on collecting a dataset that is suitable for learning new tasks offline, relying only on small amounts of task-specific online experience \citep[e.g.][]{fakoor2019metaq,cabi2020scaling,siegel2020keep,singh2020cog}. And in a similar vein we may use historic experience to mitigate problems associated with catastrophic forgetting \citep[e.g.][]{rolnick2018experience}.
Many of these ideas are already present in the literature. However, we believe that embracing the versatility of off-policy learning and a stricter separation between data collection and inference will lead to future gains:

Model-free off-policy algorithms have improved considerably and are now widely used \citep[e.g][]{heess2015learning,lillicrap2016continuous,abdolmaleki2018maximum,haarnoja2018soft}. However, they often continue to operate in an online fashion, without a clear separation of policy optimization and data collection. A more recent development are specialized algorithms that successfully  operate in fully offline settings where a policy is optimized from a fixed dataset without further interaction with the environment  \cite{fujimoto2018offpolicy,siegel2020keep,wang2020critic,abdolmaleki2021multi}. Moving forward it will be important to focus on algorithms that work well in \emph{both} the online and the offline setting \citep[e.g.][]{jeong2020learning,peng2020advantage} and can, for instance, combine large, stored 'offline' datasets and smaller amounts of 'online' experience.

The separation between the behavior that is executed and behavior that is optimized during learning has been exploited in goal-conditional or multi-task settings \citep[e.g.][]{andrychowicz2017hindsight,kalashnikov2021mtopt} and hierarchical goal-conditional \citep[e.g.][]{nachum2018data,levy2018hierarchical} settings.
More generally, there is growing interest in off-policy HRL algorithms \citep[e.g.][]{nachum2018data,galashov2018information,nachum2019does,tirumala2020behavior,wulfmeier2020compositional,wulfmeier2020dataefficient} and offline skill-learning architectures \citep[e.g.][]{krishnan2017discovery,wang2017robust,MerelNPMP,lynch2019learning,ajay2021opal,chebotar2021actionable}. So far, online and offline skill-learning architectures have, however, remained largely disjoint, and skills tend to be reused as part of a hierarchical target policy.
The \CnI~paradigm encourages us to consider further integration of online- and offline skill learning architectures, a shift from the use of skills for policy optimization towards data collection, and more generally novel synergies between transfer via experience (data) and via parameterized representations \cite[e.g.][]{cabi2020scaling,siegel2020keep,singh2020cog,ajay2021opal}.

Similar to skills, stored trajectory data can be used to learn dynamics models for model based policy optimization \citep[e.g.][]{heess2015learning,gu2016continuous,byravan2019imagined,hafner2019dream}. The \CnI~paradigm suggests instead, to use such models for online behavior optimization during data collection. This can have the benefit that model error does not directly affect the learned policy  \citep[e.g.][]{springenberg2020local,schrittwieserlearned,piche2018probabilistic,lowrey2018plan}, and that behavior can adapt rapidly, e.g.\ to alternative rewards. More generally, the \CnI~paradigm naturally allows for multiple behavior rules with different levels of amortization/'on-the-fly' optimization (see also e.g.\ \cite{jeong2020learning,galashov2020importance}). 

The \CnI~ viewpoint emphasizes the optimization of data collection. While a fully Bayesian treatment can provide an optimal trade-off between exploration and exploitation \cite{vlassis2012bayesian,ghavamazadeh2016bayesian} but it is usually intractable. Various alternative objectives have been explored both in the supervised and unsupervised setting. These include approximate treatments of uncertainty \citep[e.g.][]{bellemare2016count,osband2019deep,BitByBit2021}, intrinsic rewards derived from sensor changes \citep[e.g.][]{hertweck2020simple,hafner2020towards}, motivated by empowerment or related information-theoretic formulations \citep[e.g.][]{gregor2016variational,florensa2017stochastic,hausman2018learning,eysenbach2018diversity} and curiosity-based objectives \cite[e.g.][]{schmidhuber2010formal,pathak2017curiosity,burda2018large,sekar2020planning}. 
The \CnI~model encourages further research into strategies for the acquisition of information, and it provides a flexible framework that may facilitate a conceptual disentanglement of objectives, representations, and execution strategies.

\section{Conclusions and outlook}
\label{sec:questions}

The central message of the \CnI-paradigm is to re-think data-efficient RL via a clear separation of data collection and exploitation into two distinct but connected processes and to exploit the flexibility of off-policy RL in agent design for problems as diverse as online RL, offline RL, or lifelong-learning. This will hopefully inspire and intensify a couple of research avenues, of which we just want to highlight a few:

\begin{itemize}
    \item An optimal collect process in the sense of O2 is key for data-efficient agents and thus for achieving AI. This requires awareness of the knowledge that the agent has already acquired. A dedicated research agenda should consider: 
    What is a good objective for collecting the 'right' data?
    What surrogates could we use if the 'correct' objective is impractical?
    \item How to implement 'infer' properly and effectively (i.e. how to squeeze all of the knowledge 
    out of existing data); how to learn efficiently and reliably from large existing datasets, and how to optimally merge the on-line and the off-line viewpoint on data generation and exploitation?
    \item  \CnI~emphasizes the reuse of previously collected experience. This raises the question what other intermediate representations of knowledge, besides policies, can be extracted from data and efficiently reintegrated into the process of data collection and inference (e.g. skills, models, rewards) to improve the capabilities of the learning system?
\end{itemize}

\CnI~is not tailored to a particular learning scenario and its applications range from
'classical' single task learning scenarios to multi-task scenarios. Going forward, we see \CnI~as a
natural basis for a data-efficient learning agent, that treats data as a raw resource that can be flexibly transformed into different types of representations that can be used, for instance, for action selection (e.g.\ policies), or may facilitate future learning problems (e.g.\ models, perceptual representations, or skills). At any given time, the agent may act to collect new data, either with the goal of improving its performance on a particular, external task, or to simply learn more about its environment in a way that can be exploited in the future.

\acknowledgments{Thanks to the Control Team and various colleagues at DeepMind for ongoing discussions, and in 
particular to Patrick Pilarski, Dan Mankowitz and Markus Wulfmeier for their valuable comments on the manuscript.}


\bibliography{example,nicolas}  

\appendix

\newpage

\section{An agent architecture for Collect and Infer \label{architecture}}

The following section gives an example, of how a prototypical \CnI{} agent architecture might look. The purpose
is to give an idea of how such an agent might be implemented; it is not meant as being the only possible way to realize the \CnI{} framework in an agent.

A prototypical \CnI{} agent architecture consists of five main components: 
an actor, one or more learners, a transition memory, 
a policy pool and a scheduler. While actors and learners are common parts of any RL agent, and transition memories are found in most agents that care for data-efficiency, policy-pool and scheduler are not necessarily standard components in a normal RL agent,  but are key ingredients of this exemplary Collect and Infer agent.

In the following, the components are described in more detail.

\begin{figure*}[htb]
	\centering
	\includegraphics[width=0.4 \columnwidth]{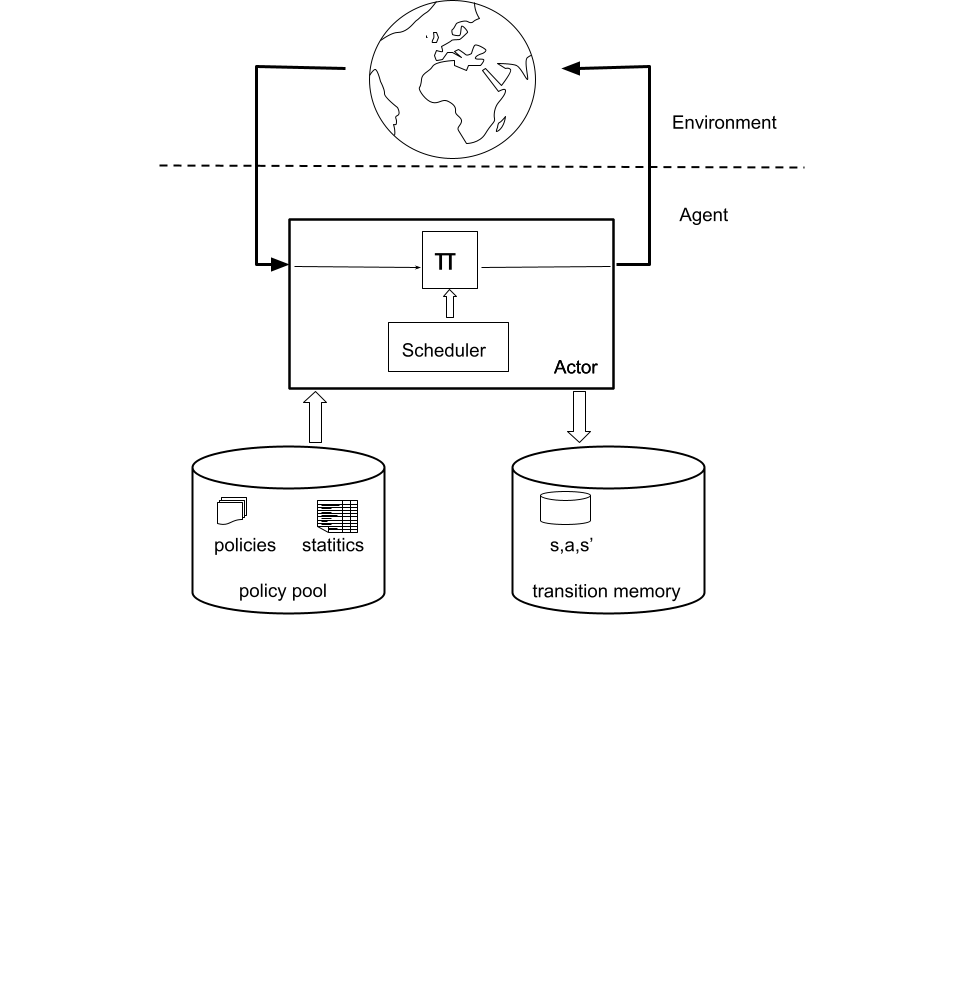}
    \includegraphics[width=0.4 \columnwidth]{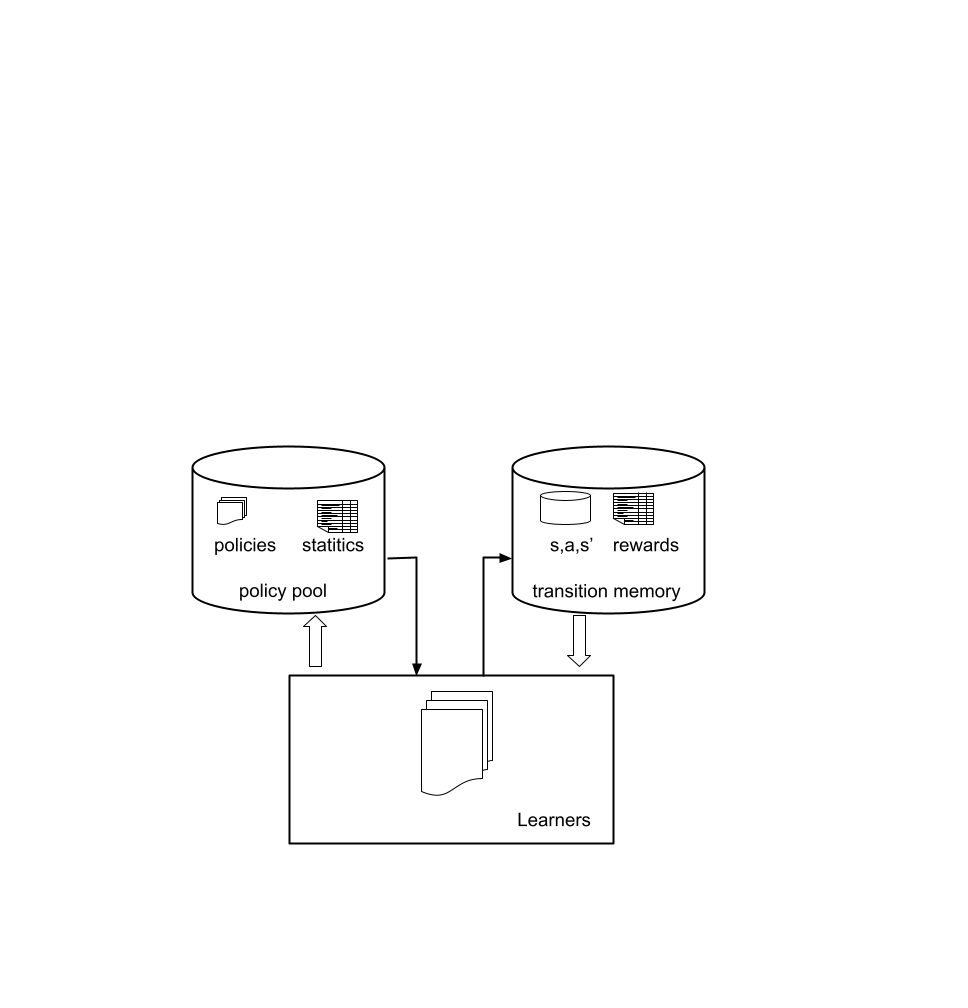}
\vspace*{-2em}	
	\caption{Parts of a \CnI{} agent. Left side shows the 'collect' part of the agent that interacts with the environment. The active policy is selected from a policy pool, transitions are observed and stored in a transition memory. Right side shows the 'infer' part of the agent: transitions are taken by the learners from the transition memory; learners produce policies and other knowledge that then is pushed back to the policy pool.}
	\label{fig:schematic2}
\end{figure*}

\subsection{The actor}

The actor implements the interface of the agent with the environment: in every time step, it receives the current observation from the environment and computes the action that is subsequently applied to the environment. The actual mapping from the observation to the action is done by the currently active policy within the actor. This active policy is selected by the  so-called '{\em scheduler}'.

\subsection{The scheduler}

The task of the scheduler is to select the currently active policy from the policy pool. In the learning phase, the scheduler aims to select the policy that promises to collect the 'right' experience, i.e. those transitions that support the learners with the most useful information.

The scheduler is also the place to implement meta-exploration strategies like multi-step actions \cite{SR03:NCAF, dabney2020temporallyextended}. It therefore is the central module to implement the
optimal collection strategy that lead to data-efficient learning.

To do so, it uses meta-information about the policies available in the policy-pool, e.g. statistics about the reward they collected thus far.

In the application phase, the role of the scheduler is to fulfill external demands by selecting the right policy that achieves the desired task. This might also be
done by selecting a sequence of policies in a hierarchical setting. Discussing this
is however beyond the intended scope of the paper.

\subsection{The learners}

In Collect and Infer, learners are generally thought of as modules that transform the information contained in the transition memory into useful pieces of knowledge. All learning therefore is inherently offline and off-policy learning, since it considers the whole batch of transition data as input. 

Learners can extract knowledge from the transition memory in many different forms:

\begin{itemize}
    \item policies (the most common outcome)
   \item environment models, e.g. for planning or exploration
   \item skills
   \item rewards
   \item observer/ representation
   \item statistics
 \end{itemize}
 
 In contrast to other RL frameworks, actors and learners are only loosely connected through the transition memory. This opens the opportunity, that learners can be thought of continuously and asynchronously 'work in the background',
 independent of the actual data collection process provided by the actor (similar to 'dreaming' in humans). The number of learners and their goals are only limited by the amount of compute resources provided to the agent.

 Learners might also consider sampling strategies to sample the most useful samples out of the transition memory. Sampling might depend on several inputs, e.g. the purpose of the learner, or the data distribution and characteristics in the transition memory (e.g. prefer transitions with positive reward).

\subsection{The policy pool}

The policy pool contains the set of policies available to the actor for the actual determination of the action.

A policy is a mapping from a stream of observations to an action. In the simplest case, it can be a function that maps the current observation to an action (e.g. a feedforward neural network). In general, a policy might include an 'observer' module, that maps the observation to a state, either explicitly by a designated module or implicitly by the use of recurrent connections or inputs with past information.
In addition, determining the action can also be a more involved process, e.g. the result of a planning process with a learned environment model. 

Policies in the pool can serve different purposes, e.g. pursue a certain task, execute a certain movement (like a motion primitive or skill in robotics), or serve exploring the environment (e.g. count-based or curiosity based exploration). The policy pool manages information about the available policies (e.g. purpose, input specification, output specification) and optionally also about their characteristics e.g. in form of statistics about collected reward. This meta-information can, for example, be used by the scheduler to determine the best policy to collect more data with.

\subsection{The transition memory}
In the spirit of 'data is precious, data is true', Collect and Infer follows the principle of storing all transition data that has been collected by the agent.
Transition data is a triple consisting of observation, action and resulting observation in the next time step. In addition, meta information like belonging to a sequence of transitions (like an episode) is recorded. In principle, the transition based memory can be seen as a sample based model of the environment.

From a practical efficiency point of view, sampling from this potentially very large memory into a smaller 'working memory' might be a reasonable or necessary step.

Conceptionally, reward is not stored with this transition. In theory, reward is only assigned at the moment, when learning for a particular purpose occurs. This makes each transition much more effective, since it can be dynamically used for learning arbitrary tasks. Of course, this does not preclude, that for computational efficiency, reward assignments are temporarily kept in memory.

\section{Case study: Collect and Infer in Robotics}

\begin{figure}[h]
	\centering
	\includegraphics[width=0.24\columnwidth]
	 {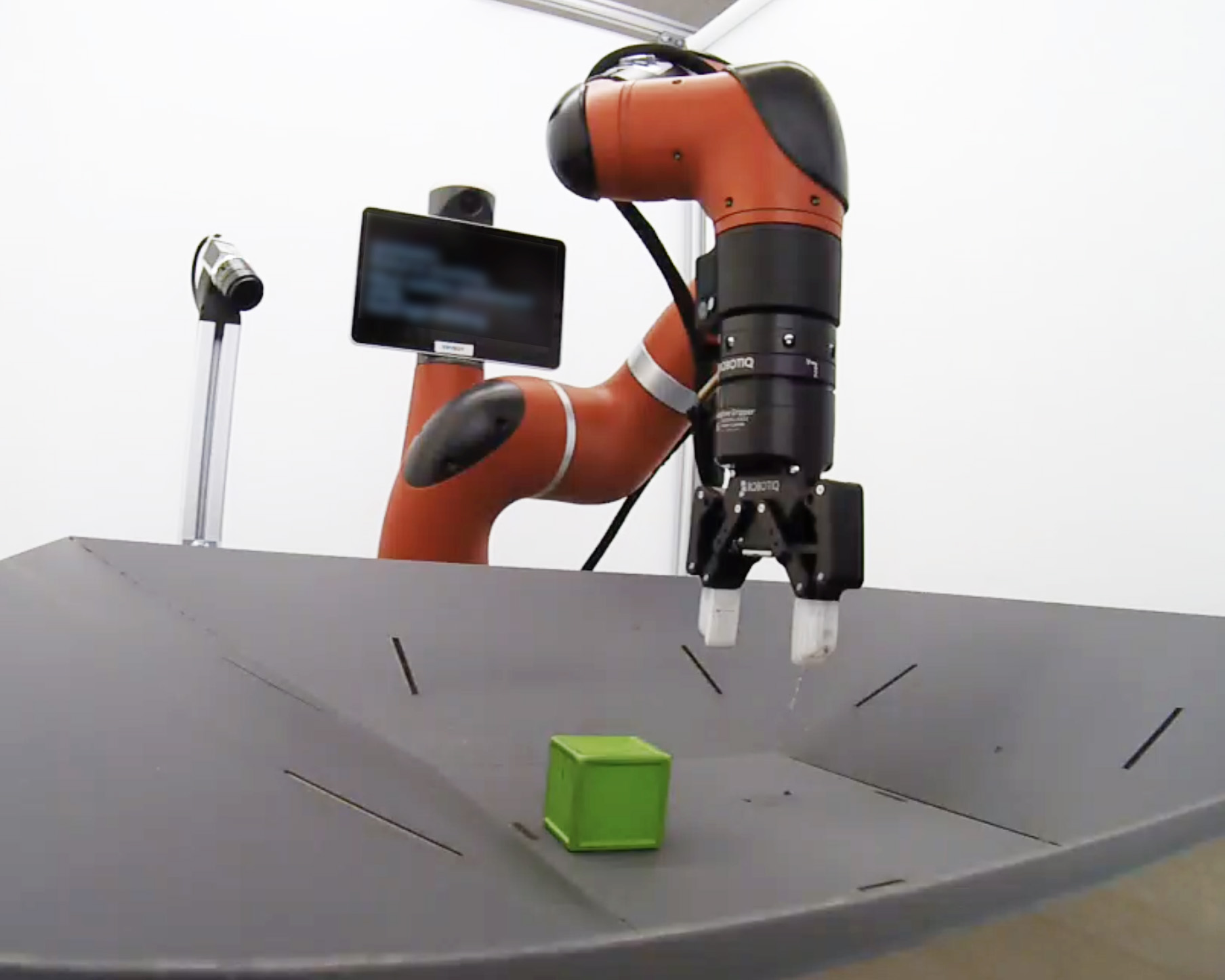}
    \includegraphics[width=0.24\columnwidth]{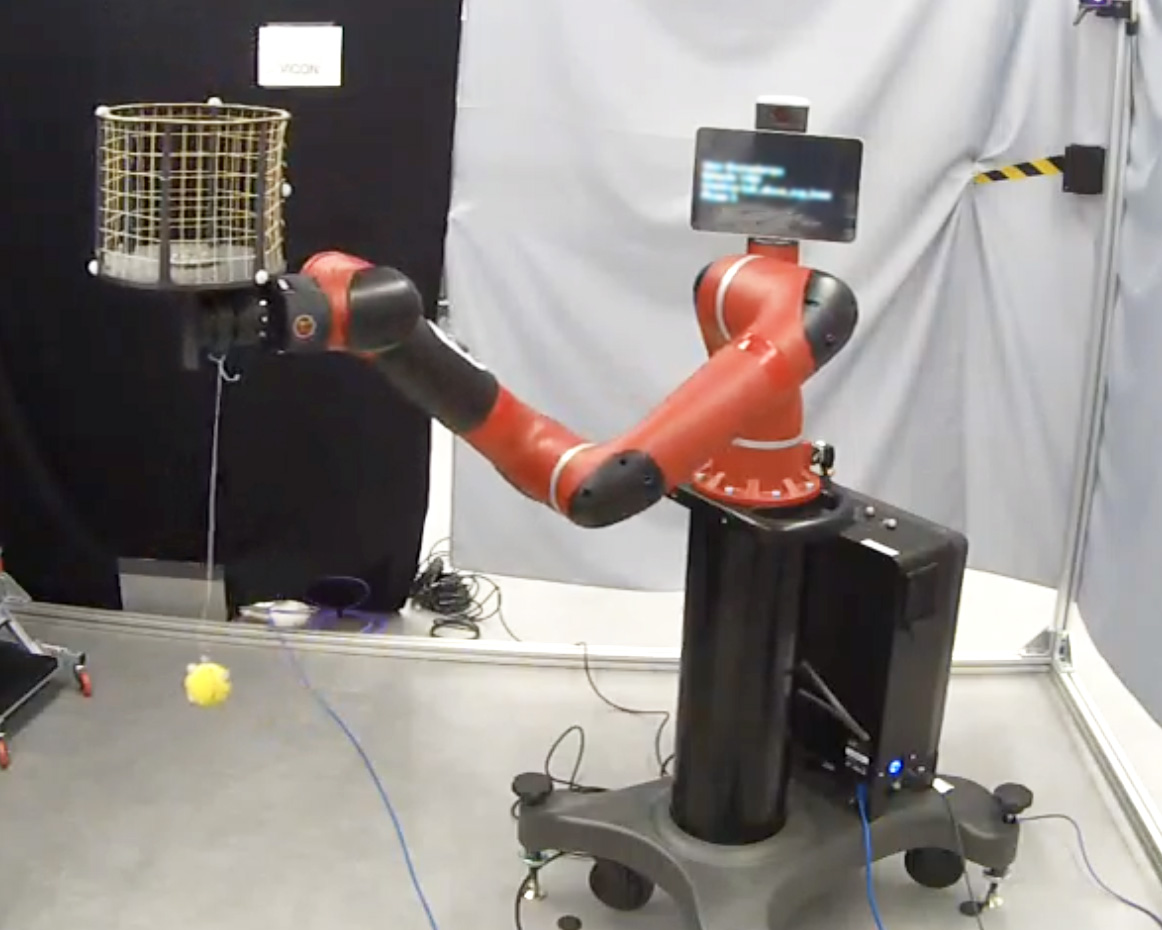}
    \includegraphics[width=0.24\columnwidth]{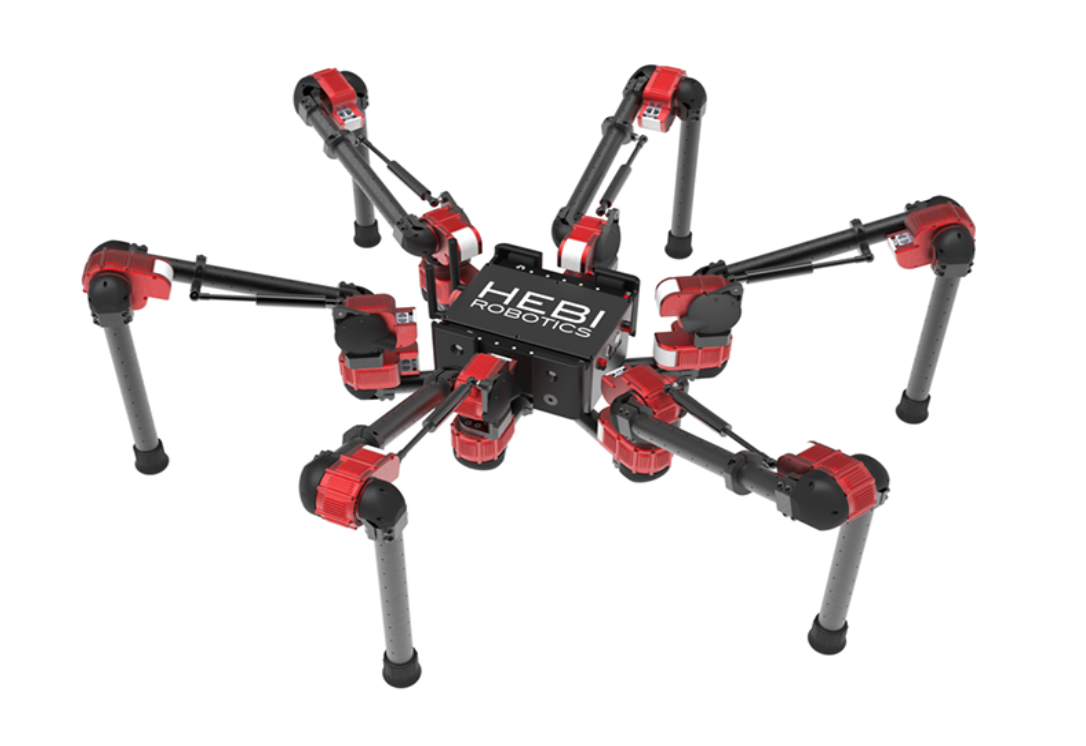}
    \includegraphics[width=0.24\columnwidth]{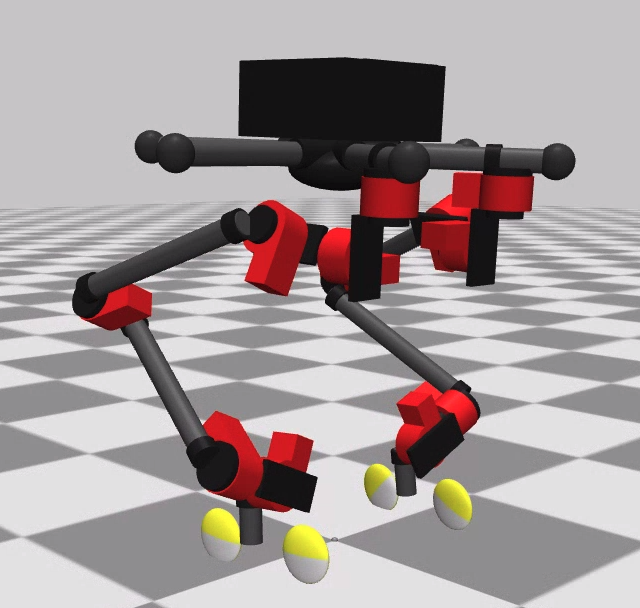}    
	\caption{\label{fig:robots_manip} The \CnI{} principle can help to develop data efficient algorithms that help to learn on real robots with minimal prior knowledge.}
\end{figure}

Solving robotic applications by Reinforcement Learning is challenging, since learning problems are
typically extremely complex due to a high-dimensional action space and considerable horizon of the 
sequential decision problem. On the other hand, data is typically limited, since it has to be collected
in the real-world via real-time interactions. Typical approaches like simulation to reality transfer make use of prior knowledge in
terms of availability of an accurate simulator. Alternatively, learning from demonstrations assumes
the existence of an already working controller from which the agent policy can be copied from.

\begin{figure}[h]
	\centering
	\includegraphics[width=0.9\columnwidth]{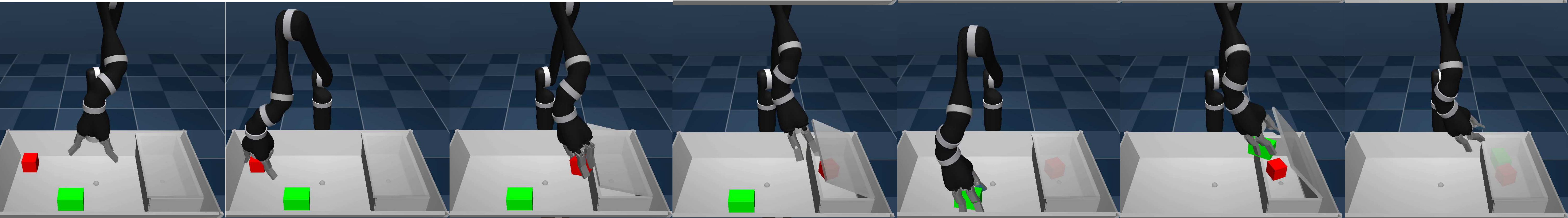}
	\caption{Example of a successful application of SAC-X, following the basic \CnI{} principles, to learn to solve a complex task of putting two objects in a box after opening the lid first. }
	\label{fig:robots_putinbox}
\end{figure}

If one wants to learn 'from scratch', data-efficient methods are key. The \CnI{} framework offers
a promising way to realize a data-efficient agent that reduces the amount of prior knowledge required by
learning directly from interactions with the real system.

The idea is to enable the agent to efficiently collect relevant transition data to finally solve a
complex target task. This can be done by specifying a set of basic auxiliary behaviours, like they are used in Scheduled Auxiliary Control (SAC-X, \cite{riedmiller2018learning}), that are considered
useful in the sense of collecting interesting and relevant data to solve the final task. As an example,
if a robot should finally learn to grasp and lift objects, a potentially useful set of behaviours is to reach
the object first, open and close the gripper, lift the arm, etc. -  a bit similar to the idea to discover the
abilities of the body. These behaviours can be specified
by auxiliary rewards, that are intrinsic to the agent. Of course, the 'right' definition of these auxiliary
rewards again requires a priori knowledge and insight. However, this kind of prior knowledge is often 
easy to define, plus, as long as the set of generated behaviours is diverse enough, the exact definition
of individual auxiliary skills does not matter too much.

Research also tries to minimize prior knowledge by defining very general auxiliary tasks, like e.g. rewards for
'modifying sensor values'. These 'simple sensor intentions' \cite{hertweck2020simple} have proven to be a powerful and general
way to 'invent' intrinsic rewards and skills, that even can learn general grasping policies or dynamic
tasks like 'ball in cup' from scratch on a real robot. A similar approach adapted for locomotion has
recently shown surprisingly data-efficient learning on all kinds of locomotion platforms \cite{hafner2020general}.

Another interesting direction is to incorporate general curiosity based reward schemes into \CnI{} based
learning architectures. Curiosity based methods are among the most general reward schemes and some first
promising results in a SAC-X based learning agents have been received recently.

A more technical advantage from realising a \CnI-perspective in an agent is the conceptual and potentially
physical separation of actor and learners: these allow the actors to run close to the real world system,
obeying real-time constraints, since their only task is to execute the current policy and to store the
transition data. Learners on the other hand can life in a physically separated space like a data-center
and potentially use a lot of computation power to update the knowledge of the agent.

\section{Some high-level inspirations from the \CnI-perspective}
\begin{enumerate}
    \item 'Data is precious and data is true': experience plays a central role. All experience is kept in memory. 
    \begin{enumerate}
        \item All knowledge at any time can be extracted from that data. 
        \item The task of the learners is to transfer that data in useful pieces of representation of this knowledge: policies, models, rewards, observers, skills.
    \end{enumerate}
    \item Exploration is key for data-efficiency: collect ‘meaningful’ information
    \item All final task policies are closed-loop and 
    \begin{enumerate}
        \item are model-free (the true model is the transition data)
        \item act on a base time-step and on a base action set 
     \end{enumerate}
    \item Since (all) environment models are never exact, ideally avoid to use models for the final task policy (e.g. as in MPC). However, environment models and planning might be useful for exploration to collect interesting data.
    \item Options/ skills might not be optimal to solve the final task. However, they are useful for exploration.
    \item State representation is key for generalisation and should be learned from data.
    \item Tasks are given externally, but rewards are in first place an agent-internal concept. Rewards are created by the agent to specify an externally given task, or to incentivize exploration of an unknown environment. 
\end{enumerate}

\section{Relation of Collect \& Infer to standard RL algorithms}
We give some insight into how the \CnI~formalism can be used to express standard RL methods, and how it might lead to interesting new derivative algorithms.

\subsection{Off-policy Q-learning as an instance of 'Collect \& Infer'}
To understand the relation of \CnI{} to classical algorithms we can first consider how standard off-policy RL can be expressed in the \CnI~ formalism described in the main paper. For this we let the infer objective $\mathcal{L}_\mathcal{I}$ consist of two parts: first we need to learn an approximation of the optimal action-value function $Q^*(s, a) = r(s) + \gamma \arg \max_{a \in \mathcal{A}} \mathbb{E}_{s' \sim p(s' | s, a)}[Q(s, a')]$. This is a particularly appealing choice as a parametric action-value function $Q_\phi(s, a)$ -- with parameters $\phi$ -- can be learned off-policy from the dataset $\mathcal{D}_c$ by maximizing the objective:
\begin{equation}
    \mathcal{L}_I(\pi_{\theta_Q}, \mathcal{D}_c) =  -\mathbb{E}_{(s_t, a_t, s_{t+1}) \sim \mathcal{D}_c} \Big[ \big(r(s) + \gamma \arg \max_{a \in \mathcal{A}} \mathbb{E}_{s' \sim p(s' | s, a)}[Q^*_{\theta_Q'}(s, a')] - Q^*_{\theta_Q}(s, a) \big)^2 \Big],
\end{equation}
where $\theta_Q'$ are the parameters of a target network (that are periodically copied from $\theta_Q$ during optimization). And where we define the policy $\pi_{\theta_Q}$ as choosing the action with maximum Q-value in each state. This leads to the deterministic policy: $$
\pi_{\theta_Q}(s) = \arg \max_{a \in \mathcal{A}} Q^*_{\theta_Q}(s, a),
$$
which can be reasonably represented if the set of actions $\mathcal{A}$ is finite. As can be seen this optimization is analogous to DQN \citep{mnih2015human}. In this case, given the knowledge that Q-learning can work on any data, the collection process $\CollectOp$ could, in theory be a completely random policy (assuming infinite dataset size for $\mathcal{D}_c$). 
Practical algorithms  tightly interleave data collection and inference. This allows more efficient data collection than with a random policy, for instance by using perturbed versions of the current optimal policy (e.g. epsilon greedy), or via more advanced strategies, that approximately take uncertainty about the Q-function into account \citep[e.g.][]{osband2019deep}. Inference is usually performed incrementally, with updates to the Q-function being applied after every environment interaction. To improve inference efficiency the replay buffer is also often limited in size. However,  \eqref{eq:collect_obj} and the ideas discussed above suggest the consideration of alternative schemes. These could involve, for instance, replacing the frequent incremental updates with less frequent inference steps during which all model parameters are fully re-estimated using all data collected so far; or the use of data collection (exploration) strategies that deviate more strongly from the current optimal policy, possibly via a meta-learned exploration strategy that can take the current knowledge state of the agent into account.

\subsection{Off-policy Actor-Critic as an instance of 'Collect \& Infer'}
The argument for Q-learning from above can be extended to the off-policy actor critic setting \citep{lillicrap2016continuous,abdolmaleki2018maximum,haarnoja2018soft} in which we consider separating Q-function and policy (e.g. because we consider continuous actions). In this case we can still learn a parametric action-value function $Q_\phi(s, a)$ -- with parameters $\phi$ -- off-policy from the dataset $\mathcal{D}_c$ by maximizing the objective:
\begin{equation}
    \mathcal{L}_Q(\phi, \mathcal{D}_c) = - \mathbb{E}_{(s_t, a_t, s_{t+1}) \sim \mathcal{D}_c} \Big[ \big(r(s) + \gamma \mathbb{E}_{s' \sim p(s' | s, a), a' \sim \phi_\theta}[Q_\phi(s, a')] - Q_\phi(s, a) \big)^2 \Big].
\end{equation}
The solution for the optimal policy then is given as the result of the maximization:
$$
\mathcal{L}_\pi(\pi_\theta, \mathcal{D}_c) =  \mathbb{E}_{\tau \sim \mathcal{D}_c} \Big[ \sum_{s \in \tau} \mathbb{E}_{a \sim \pi_\theta} \big[ Q_\phi(s, a) \big] \Big],
$$
where we consider a stochastic policy $\pi_\theta(a | s)$. Taken together we can express the infer objective as 
$$
\mathcal{L}_\mathcal{I}(\Theta, \mathcal{D}_c) = \mathcal{L}_Q(\phi, \mathcal{D}_c) + \mathcal{L}_\pi(\pi_\theta, \mathcal{D}_c),
$$
where $\Theta = \lbrace \phi, \theta \rbrace$ are the combined parameter of policy and Q-function. As in the example for standard Q-learning above, we can derive an optimal collection policy for this inference process, and doing so would lead to new algorithms.

\subsection{Behavior Cloning with an optimal teacher as an instance of 'Collect \& Infer'}
Another interesting choice for an inference objective is to consider a distance function around the action that was actually executed when collecting the trajectory $\pi$. Assuming a deterministic policy, this leads to an objective for the infer part of learning that is analogous to supervised learning or behavior cloning:
$$
\mathcal{L}_\mathcal{I}(\pi_\theta, \mathcal{D}_c) =  -\mathbb{E}_{\tau \sim \mathcal{D}_c} \Big[ \sum_{s, a \in \tau}  \big(a - \pi_\theta(s) \big)^2 \Big],
$$
where we chose a squared distance function and a deterministic policy $a = \pi_\theta(s)$. In this scenario it is then interesting to consider what the optimal collection policy looks like. It is intuitive that the policy maximizing the performance of $\pi_\theta$ in this case corresponds to the optimal deterministic policy $\pi^*(s)$ itself\footnote{We refer to \citet{rajaraman20} for a recent discussion of optimality conditions for behavior cloning.}, that is 
$$
\arg \max_{\pi_c} \mathcal{O}(\mathcal{C}, \mathcal{I}, N) = \arg \max_{\pi_c} J(\pi_c), \text{ where } \mathcal{D}_c = \lbrace \tau^1, \dots, \tau^N | \tau^i \sim \pi_c, p \rbrace = \mathcal{C}(\mathcal{E}, N).
$$
And thus, in this case, the collection of the data itself becomes equivalent to the RL problem of finding an optimal policy -- while inference is performed merely by supervised learning.

\end{document}